\documentclass[tinyml]{acmart}

\AtBeginDocument{%
  \providecommand\BibTeX{{%
    \normalfont B\kern-0.5em{\scshape i\kern-0.25em b}\kern-0.8em\TeX}}}

\setcopyright{rightsretained}
\copyrightyear{2024}
\acmYear{2024}
\usepackage[utf8]{inputenc} 
\usepackage[T1]{fontenc}    
\usepackage{hyperref}       
\usepackage{url}            
\usepackage{booktabs}       
\usepackage{amsfonts}       
\usepackage{nicefrac}       
\usepackage{microtype}      
\usepackage{xcolor}         
\usepackage{microtype} 
\usepackage{booktabs}  
\usepackage{url}  

\usepackage{amsmath}
\usepackage{amsthm}
\usepackage{xspace}
\usepackage{caption}
\usepackage{multirow}
\usepackage{booktabs}
\usepackage{natbib}
\usepackage[export]{adjustbox}
\usepackage{placeins}
\usepackage{algorithm}
\usepackage{subfigure}
\usepackage{algpseudocode}
\usepackage{bbm}
\usepackage{wrapfig}
\usepackage{lipsum}
\usepackage[T1]{fontenc}       \usepackage{booktabs}
\usepackage[T1]{fontenc}
\usepackage{amsmath, amsfonts}
\usepackage{ifthen}
\usepackage{booktabs}
\usepackage{microtype}
\usepackage{placeins}
\usepackage{graphicx}
\usepackage{xspace}

\usepackage[detect-weight=true, detect-family=true]{siunitx}
\usepackage[colorinlistoftodos,linecolor=blue,backgroundcolor=white,bordercolor=blue,textsize=tiny,textwidth=18mm]
{todonotes}
\usepackage[utf8]{inputenc}
\usepackage[detect-weight=true, detect-family=true]{siunitx}
\DeclareRobustCommand{\rchi}{{\mathpalette\irchi\relax}}
\newcommand{\irchi}[2]{\raisebox{\depth}{$#1\chi$}}
\newcommand{\sL}{\mathcal{L}}
\newcommand{\round}[1]{\num[round-mode=places,round-precision=2]{#1}}
\newcommand{\acc}[2]{\round{#1}\ifthenelse{\equal{#2}{}}{}{\tiny ${\scriptstyle \pm}$\round{#2}}}
\newcommand{\ouralg}{ScheduledKD-LDC\xspace}
\newcommand{\bacc}[2]{\bf \round{#1}\ifthenelse{\equal{#2}{}}{}{\bf \tiny ${\scriptstyle \pm}$\round{#2}}}

\usepackage{tikz}

\usepackage{mathtools}

\begin{document}

\title{Scheduled Knowledge Acquisition on Lightweight Vector Symbolic Architectures for Brain-Computer Interfaces}


\author{Yejia Liu\texorpdfstring{\textsuperscript{1}}{1}, Shijin Duan\texorpdfstring{\textsuperscript{2}}{2}, Xiaolin Xu\texorpdfstring{\textsuperscript{2}}{2}, and Shaolei Ren\texorpdfstring{\textsuperscript{1}}{1}}
\affiliation{
  \institution{\texorpdfstring{\textsuperscript{1}}{1}UC Riverside}
  \institution{\texorpdfstring{\textsuperscript{2}}{2}Northeastern University}
}
\email{{yliu807@ucr.edu}, {duan.s@northeastern.edu}, {x.xu@northeastern.edu}, {sren@ece.ucr.edu}}

\renewcommand{\shortauthors}{Trovato and Tobin, et al.}

\begin{abstract}
Brain-Computer interfaces (BCIs) are typically designed to be lightweight and responsive in real-time to provide users timely feedback. Classical feature engineering is computationally efficient but has low accuracy, whereas the recent neural networks (DNNs) improve accuracy but are computationally expensive and incur high latency. As a promising alternative, the low-dimensional computing (LDC) classifier based on vector symbolic architecture (VSA), achieves small model size yet higher accuracy than classical feature engineering methods. However, its accuracy still lags behind that of modern DNNs, making it challenging to process complex brain signals.
To improve the accuracy of a small model, knowledge distillation is a popular method. However, maintaining a constant level of distillation between the teacher and student models may not be the best way for a growing student during its progressive learning stages.
In this work, we propose a simple scheduled knowledge distillation method based on curriculum data order to enable the student to gradually build knowledge from the teacher model, controlled by an $\alpha$ scheduler. Meanwhile, we employ the LDC/VSA as the student model to enhance the on-device inference efficiency for tiny BCI devices that demand low latency. The empirical results have demonstrated that our approach achieves better tradeoff between accuracy and hardware efficiency compared to other methods. 
\end{abstract}


\maketitle

\section{Introduction}
A brain-computer interface (BCI) allows for direct communication between the human brain and an external device without the need for physical movement~\cite{bci1, bci2, bci3, guger2021brain}. The Electroencephalogram (EEG), as a typically non-invasive neuroimaging technique to measure and record the electrical activity of the brain, has been widely used in the BCI applications~\cite{wagh2022evaluating, Lawhern2018}. 
The deep neural networks (DNNs) have shown promising results in extracting spatial-temporal dynamics from EEG signals, and improved classification accuracy compared to the classical feature engineering methods~\cite{roy2019deep, supratak2017deepsleepnet, zhang2018cascade}. However, the intensive computation required by DNNs in inference can result in high latency in real-time EEG-based BCIs, which are intended to be lightweight~\cite{bci-u, nakanishi2019facilitating, zhang2020fracbnn}. The latency, for example, can pose a challenge for disabled persons using BCI-controlled prosthetic limbs, making it difficult to perform fine motor tasks such as picking up small objects or typing on a keyboard. Moreover, in some implantable BCI devices, power constraints are even more stringent, with the power needing to stay under 15-40mW to comply with FDA, FCC, and IEEE guidelines~\cite{hshalo, guide}, which immediately rules out many conventional DNN-based classifiers. 

Due to the hardware efficiency with massive processing parallelism, the binary hyperdimensional computing based on the vector symbolic architecture (HDC/VSA) stands out in the edge inference paradigm~\cite{kleyko2023survey, kleyko2022vector, kanerva2009hyperdimensional, neubert2019introduction}. In the binary HDC/VSA, objects are encoded into long, binary vectors in a high dimensional space. By exploiting the algebraic properties of a small set of operators, the observed instance vectors belonging to the same class are aggregated, allowing for the creation of a meaningful symbolic class representation in the end, for inference of new objects~\cite{kanerva2009hyperdimensional, frontier}. Although lightweight, the binary HDC/VSA suffers from low accuracy and large model size as a result of its rudimentary training procedure and high dimensionality of vectors~\cite{duan2022braininspired, duan2022lehdc, HDCSurveyReviewIEEECircuitMagazing20209107175}. The recent proposed low-dimensional computing (LDC)~\cite{duan2022braininspired} alleviates these issues by utilizing a partially binary neural network (BNN)~\cite{qin2020binary} to hash samples into binary codes with dimensionality less than $100$. Featuring a systematic training based on backpropagation, the accuracy of LDC is improved with $100\times$ smaller model size, along with faster inference speed, compared to the HDC. Nonetheless, the accuracy gap between LDC and a modern neural network still remains as a challenge, preventing its adoption in the BCIs, which involve complex neural signals as the input. 

One popular technique to improve the accuracy of a small model is knowledge distillation (KD), where a larger ``teacher'' network supervises a smaller ``student'' model, where the knowledge is distilled by the soft probabilities with a (usually static) hyperparameter $\alpha$ to balance the distillation term and the classification term~\cite{hinton2015distilling}.
However, the distillation is not always advantageous, particularly when there is a large capacity gap between big teachers and small students~\cite{cho2019efficacy, Zhao-2022-CVPR}. Drawing inspiration from curriculum learning~\cite{Bengio_curriculumlearning}, ~\cite{xiang2020learning, zhu-etal-2021-combining-curriculum-learning}, in which the student is gradually trained using samples ordered in an easy-to-hard manner. The recent CTKD curriculums the distillation temperature by adversarial training to guide the learning process for the student~\cite{li2022cur}. Although demonstrated as effective,  their training requires meticulous planning and they have not explored how to efficiently transfer knowledge to a much smaller architecture like binary LDC.

\begin{figure}[t]
\centering
\includegraphics[width=0.4\textwidth]{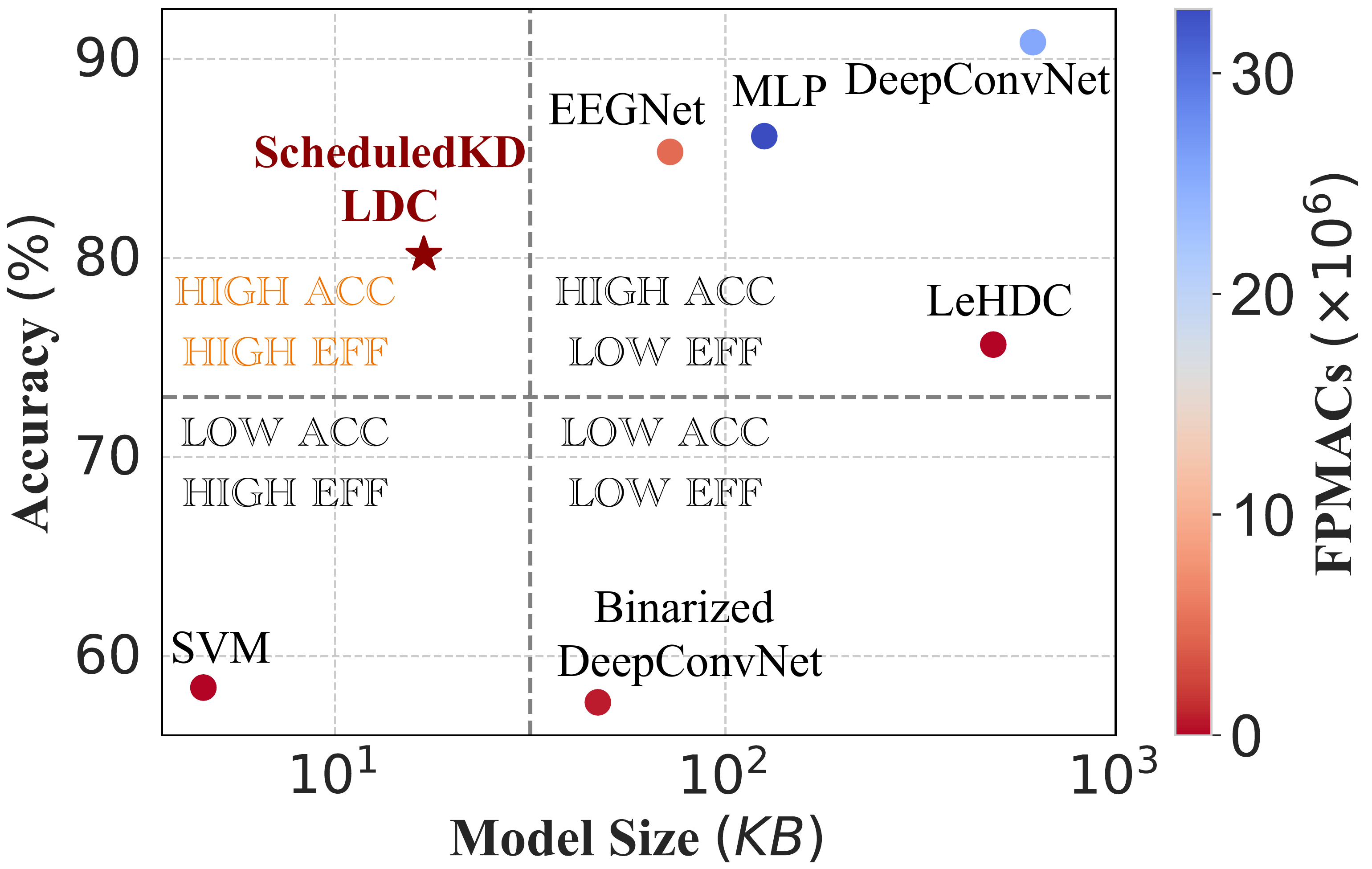}
\caption{Comparison of accuracy and inference efficiency of various methods on the Motor Imagery dataset. The \ouralg has achieved a good tradeoff between accuracy and efficiency, considering the model size and the number of FPMAC operations.
}
\label{fig:teaser}
\end{figure}

In this paper, we propose a simple yet effective approach to control the procedure of knowledge distillation from a complex teacher network to a lightweight LDC for EEG-based BCIs. We refer to the proposed approach as \textbf{\ouralg}, which uses an $\alpha$ scheduler that decreases exponentially during the distillation process, with curriculum data order. Intuitively, as the students gradually build up knowledge by learning from the confident predictions on simpler data points from the teacher, they could rely less on the teacher over time. Meanwhile, more independent learning at the later stage enhances the student model to develop its own data representations to generalize to unseen examples. As shown in Figure~\ref{fig:teaser}, the \ouralg achieves a good balance between accuracy and efficiency compared to other methods. Our empirical results indicate that it consistently outperforms other methods on the evaluated EEG datasets. 
As an east-to-use method, we believe the \ouralg is a favorable option to realize efficient edge intelligence for real-time BCIs.

\section{Preliminaries}
\begin{figure*}[t]
\centering
\includegraphics[width=0.95\textwidth, height=0.3\textheight]{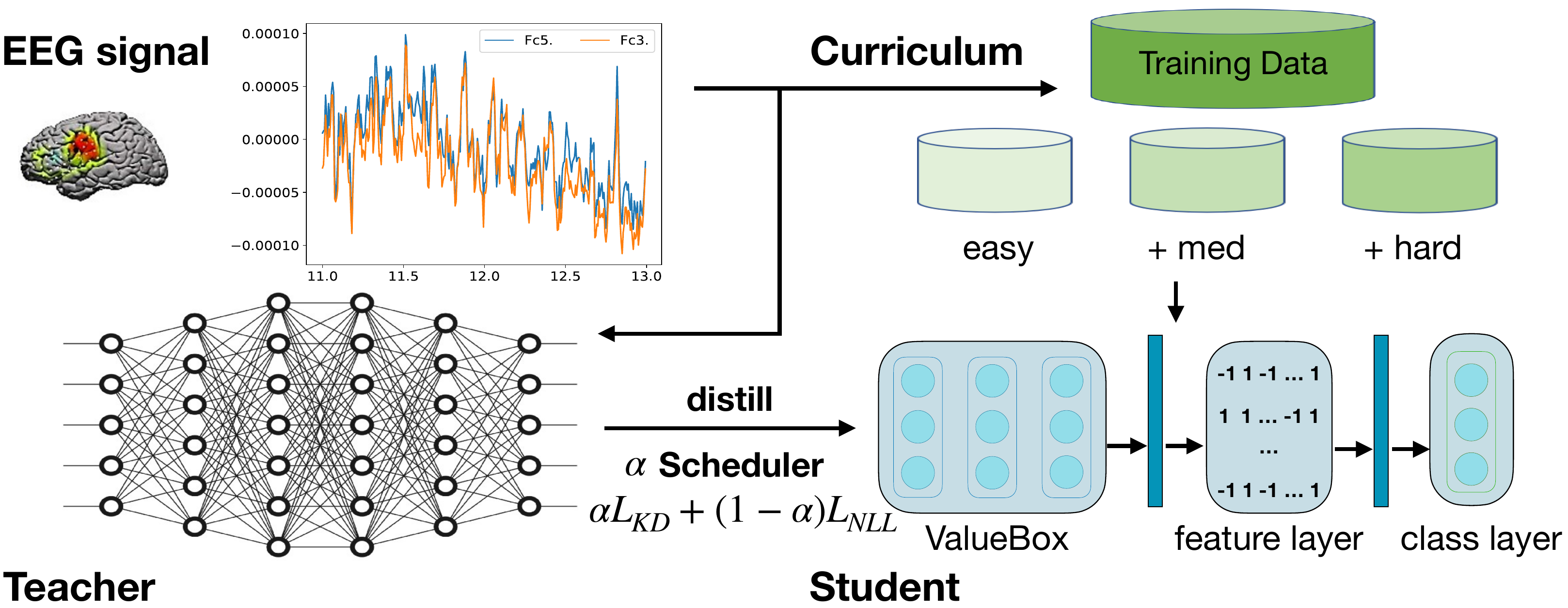}
\vspace{-6pt}
\captionsetup{width=\textwidth}
\caption{Overview of the \ouralg. We use an $\alpha$ scheduler and curriculum data order to enable the student model to gradually distill knowledge from the complex teacher model. The student model, low-dimensional computing classifier (LDC), is based on the vector symbolic architecture for efficient on-device inference.
}
\label{fig:main}
\vspace{-10pt}
\end{figure*}
We define the input space as $\rchi$ and the label space as $Y$, where
$|Y| = C$. Let $f: \rchi \times \Theta \rightarrow R^C$ be a classifier, parameterized by $\theta \in \Theta$. It outputs a categorical predictive distribution over $Y$, $\hat p(y = i | \mathbf x) = \sigma_i (f(\mathbf x, \theta))$, where $\sigma_i (\mathbf{z}) := \exp(z_i) / \sum_j \exp(z_j)$ represents the softmax function, and $z \coloneqq f(\mathbf x, \theta)$ is referred to as the logits. For simplicity, we use $f^t$ to denote the teacher model and $f^s$ to represent the student model. Knowledge distillation~\citep{hinton2015distilling} uses the soft output (logits) of one or multiple
large models as the teacher and transfers the knowledge
to a small student model, where the student model minimizes a combination of two loss objectives as shown in the Eq.(\ref{eq:loss}),
\begin{align}
    \sL := \alpha \sL_{KD} (\mathbf z^s, \mathbf z^t) + (1-\alpha) \sL_{NLL} (\mathbf z^s, y),
\label{eq:loss}
\end{align}
where $\sL_{KD}$ is the distillation term to encourage the student model to resemble teacher's responses for data examples, while $\sL_{NLL}$ is the normal cross-entropy loss between the logits $\mathbf z^s$ and the label $y$.  These two terms are balanced by the $\alpha$, which is a static hyperparameter in most existing works~\citep{Yang-2022-CVPR, Yuan2020ReinforcedMS,chen2019online, fakoor2020fast}.

Inspired by the significance of arranging information in human learning process, the curriculum training involves regulating the data order to adjust a model's learning trajectory~\citep{CL, soviany2021curriculum, liu2022nav}. A scoring function $t$ is used to determine the difficulty of each data example. If $ t(\mathbf{x_j}, y_j) > t(\mathbf{x_i}, y_i)$ for two data samples $\mathbf{x_i}$ and $\mathbf{x_j}$, we would say $x_j$ is more difficult than $x_i$. In this work, we consider the real-valued loss of a reference model that is trained on the same set of data points as our scoring function, $\mathbf t (\mathbf{x_i}, y_i) =  \ell (f_\theta(\mathbf{x_i}, y_i)) $. In essence,
presenting the data in an easy-to-hard sequence is known as \textit{curriculum} data order, while 
we use the term \textit{anti-curriculum} to describe the ranking of data from difficult to easy, and  explicitly referring to a random order as \textit{random} in this work.

\section{Student: Low-dimensional Vector Symbolic Architecture} 
In the \textbf{architecture level}, we employ the low-dimensional classifier (LDC) based on the vector symbolic architecture (VSA) as the student model for the BCI applications which demand low latency.

\subsection{Hyperdimensional Computing/Vector Symbolic Architecture (HDC/VSA)} 
The HDC/VSA represents symbolic concepts using high-dimensional distributed vectors that coexist in a shared space, providing context for each other~\cite{frontier, yu2023understanding}.
To achieve efficient hardware implementation, binary-valued hypervectors composed of $\{-1, 1\}^D$ are used, where $D$ is the vector dimension. In addition, the architecture employs a small predefined set of operators on these hypervectors with high processing parallelism, such as Binding $\bigotimes$
\footnote{e.g., $(v \otimes w) \in \{-1, 1\}^D$ and $(v \otimes w)_i = v_i w_i$.} and Bundling $\bigoplus$\footnote{e.g., $(\oplus_{k=1}^{m} w_k)_d = \sum_{k=1}^{m} (w_k)_d$ for all $d \in \{1, 2, ..., D\}$.}, 
which can be implemented by simple hardware circuits such as AND, OR gates and adders, as explained in~\cite{karunaratne2020inmemory, bi-hdc1, bi-hdc2}. 

In a standard classification task using the HDC/VSA, instances are \emph{hashed} into long binary vectors. As an example, an image $\mathbf x_i$ with features $F = [f_1, f_2, \ldots, f_N]$, and each feature with discrete values $V_{f_{i}} = [v_1, v_2, \ldots, v_M]$ can be constructed symbolically. For instance, an image in the MNIST~\cite{mnist} dataset consisting of 784 pixels, where each pixel contains 256 shades of gray, can result in $N=784$ and $M=256$. A data example $x_i$ is expressed by $\mathbf x_i := \text{sgn} ({\bigoplus_{j=1}^N f_{j}} {\bigotimes} V_{f_{j}})$ with a final dimensionality of $256$. During training, vectors from the same class are combined to create a symbolic class representation. During inference, a test image undergoes the same hashing process, and then the Hamming distance between the encoded vector and each class representation is calculated to determine the classification result~\cite{bi-hdc1, bi-hdc2, kanerva2009hyperdimensional}.

\subsection{Low-dimensional Computing (LDC) Classifier} 
The HDC/VSA blends the benefits of connectionist distributed representation and structured symbolic representation, where the representations can be composed, probed, and transformed by a set of hardware-efficent math operations~\cite{frontier, kamienny2023deep}. However, the HDC/VSA architecture has two main issues: the \textit{large model size} brought by the high dimensionality of vectors, where $D$ is typically on the order of $10,000$
; and the \textit{low accuracy} due to the simple heuristic hashing process used to generate hypervectors~\cite{duan2022lehdc, HDCSurveyReviewIEEECircuitMagazing20209107175}. To address these problems, the low-dimensional computing classifier (LDC) has been proposed~\cite{duan2022braininspired, liu2023metaldc}. 

As shown in the Student Model of Figure~\ref{fig:main}, the LDC uses a partially \textit{binary neural network} in its ValueBox which encodes the input to low-dimensional vectors through deep hashing to generate trainable $F$ and $V_{f_i}$ for a sample. Additionally, the LDC allows for the feature vector dimension $D_{f_i}$ to be a multiple of the value dimension $D_{V_{f_i}}$, enabling dimensionality reduction.
The empirical results in ~\cite{duan2022braininspired} have shown that the LDC classifier with $D < 100$ can still achieve comparable or even better accuracy than the HDC.

\section{Scheduling the Knowledge Distillation: \ouralg}
It can be challenging for a student model to replicate the predictions $\mathbf z^t$ of a teacher model when there is a significant disparity in their model capacities~\citep{cho2019efficacy, Zhao-2022-CVPR, stanton2021does}. This presents a particular problem in the context of small brain-computer interface (BCI) devices, where the employed student model like LDC has limited model size and computational complexity considering fast inference. To mitigate this issue, we propose a simple yet effective approach called the \ouralg. It integrates both an $\alpha$ scheduler, which schedules the $\alpha$ in Eq.~(\ref{eq:loss}) and a data curriculum to jointly regulate the distillation process. 
An overview of the \ouralg is provided in the Figure~\ref{fig:main}.

\textbf{Data Level} Given the limited computation and data representation capability of the lightweight LDC, presenting examples in a curated order can help in building representations step-by-step, starting from simpler concepts and gradually incorporating more intricate ones to capture complex dynamics in BCI datasets. We therefore adopt a \textit{curriculum data ordering} strategy in our approach, where we divide the training data into three pools being \{Easy\}, \{Easy, Medium Hard\} and \{Easy, Medium Hard, Hardest\}. The higher percentage allocation of training data to the easy pool (e.g. $[60\%, 70\%]$ in the experiments), allows the model to establish a strong initialization by assimilating foundational knowledge from the teacher. Conversely, the hard examples, which are usually more challenging or represent edge cases, receive a smaller percentage allocation (e.g. $[10\%, 20\%]$), as the hardest pool.

\textbf{Learning Procedure} However, the curriculum data order alone is not sufficient to address the problem of the student model struggling to learn from more complex teacher models effectively.
We therefore also introduce an $\alpha$ scheduler to \textit{manage the distilling level} from the teacher to the student model. Initially, a higher value of $\alpha$ is used to emphasize the influence of the teacher, 
as the teacher model can provide more accurate and reliable predictions on the easier examples, which allows the student model to build up its knowledge base by learning from the teacher's confident predictions. As the student model becomes more sophisticated, we decrease the value of $\alpha$ to let it learn more by itself. The rationale of using an $\alpha$ scheduler is two-fold: 1) the small student model cannot comprehend the entire knowledge of the much more complex teacher net~\cite{cho2019efficacy}; 2) some hard examples may challenge even the teacher model, where the wrong prediction can lead to poor performance if using the same level of distillation strength.
By allowing the more mature student model to learn from itself, it can develop its own data representations that may generalize better to unseen examples. 

When it comes to the choice of the $\alpha$ scheduler, we consider several desirable properties. First is the \textit{gradual transition}. A gradual shift helps prevent sudden disruptions and provides a more stable learning process for the student model. Secondly, \textit{flexibility} is important to cater to different tasks and dataset. For example, some tasks may require longer guidance from the teacher, while others may benefit from faster independence of the student. Thus, the scheduler should allow customization to accommodate these varying needs. We therefore propose using the exponential scheduler due to its smoother change compared to the linear one, as well as its customization capability compared to a static $\alpha$. Specifically, after the  $P$th epoch, the value of $\alpha$ is exponentially decreased by $\gamma^{\lceil \frac{h}{r} \rceil}$, where $\gamma$ is the decay rate, $h$ represents the epoch number and $r$ is the scaling factor. Additionally, our empirical results demonstrate that using the exponential $\alpha$ scheduler yields higher accuracy than directly optimizing $\alpha$ by parameterizing it, as the difference in magnitudes between the values of $\sL_{KD}$ and $\sL_{NLL}$ heavily biases the training process towards prioritizing one loss over the other, leading to worse performance. It is also worth mentioning that for more challenging data, smaller values of $P$ in general leads to a slightly higher accuracy through our observation.
We outline our algorithm in Algorithm~\ref{algo:skd}.

\begin{algorithm}[t]
\small
  \caption{Scheduled Knowledge Distillation}
  \label{algo:skd}
\begin{algorithmic}
   \State {\bfseries Input:} Training data $\{x_i, y_i\}_{i=1}^{I}$; Total training epoch $H$; Pretrained teacher model $f^t$ with $\theta_{t}$; Student model $f^s$ with randomly initialized $\theta_{s}$; Initial balancing weight $\alpha$; Difficulty ranking function $t$; Decay step $k$; Decay rate $\gamma$; Change point $P$; Order $o \in$ \{``curriculum", ``random", ``anti-curriculum"\}.
   \State {\bfseries Output:} Trained student model $f^s$.
   \State {Rank data:} $(\mathbf x_1, \mathbf x_2, ..., \mathbf x_I) \leftarrow \text{sort}({\mathbf x_1, \mathbf x_2, ..., \mathbf x_I}, s, o)$
   \While {$h < H$}
    \If {$h \geq P$ and $h \, \% \, k = 0$}
    \State Exponentially decrease $\alpha$: $\alpha \leftarrow \alpha \times \gamma^{\lceil \frac{h}{r} \rceil}$
    \EndIf
   \State Update $\theta_s$ based on the curriculum: $\theta_{s}^{h} \leftarrow \text{train-one-epoch}(\theta_s^{h-1}, \{\mathbf x_1, \mathbf x_2, ..., \mathbf x_I\})$
  
   \EndWhile

   \end{algorithmic}

\end{algorithm}

\section{Experiments}
\subsection{Experimental Setup}
\label{sSetup}
\textbf{Datasets}
we have considered three EEG datasets in experiments. The Motor Imagery EEG signals were recorded from participants who were instructed to perform or imagine performing one of five movements: eyes closed, both feet, both fists, left fist, and right fist~\citep{motor}. 
The X11 and S4b datasets from the BCI competition IIIb benchmark~\citep{939829} involves classifying left- and right- hand movements based on EEG signals.
The ERN dataset from~\cite{ern} includes 26 participants completing a P300 speller task. 

\textbf{Evaluation Metrics}
In the evaluation, we take into account both the efficiency and accuracy. Specifically, we assess the binary multiply-accumulate (BMAC) operations~\cite{zhang2020fracbnn}, floating point multiply-accumulate (FPMAC) operations~\cite{basiri2014efficient}, and model size required for inference to determine the efficiency of evaluated methods. The BMAC operations can be executed using XNORs and population counts (popcnt) in a highly hardware-efficient manner~\cite{zhang2020fracbnn}.
Therefore, models dominated by BMACs typically exhibit significantly improved inference speed and reduced model size compared to the ones governed by FPMACs~\citep{duan2022braininspired, zhang2020fracbnn}.

\textbf{Baselines}
The EEGNet is a compact deep neural network specially designed for EEG-based tasks~\citep{Lawhern2018}, 
Our experimental DeepConvNet is composed of five convolutional layers,
followed by a fully connected layer. 
In addition, we have binarized the conv layers of these two deep networks for a more comprehensive comparison. These binarized models were respectively named the Binarized-DeepConvNet and the Binarized-EEGNet. Also, we considered the Multi-Layer Perceptron (MLP) 
and the Support Vector Machines (SVM) in our experiments. The LeHDC is based on the HDC/VSA architecture but trains it with a systematic learning strategy~\citep{duan2022lehdc}. 
Besides the standard knowledge distillation method for the LDC (KD-LDC), we also have compared with
the CTKD~\citep{li2022cur}, where the curriculum is applied on the distilling temperature by an adversarial manner to control the information transfer.

\textbf{Implementation Details} 
During training, we set the learning rate for the Motor Imagery dataset to $0.005$, decaying by a factor of $0.1$ every $50$ steps, with a batch size of $1000$. For the X11 and S4b datasets, the learning rate remains at $0.005$, with a step size of $60$ and a batch size of $256$. For the LeHDC model, we set the feature dimension $D_{f_i}$ and value dimension $D_{V_{f_i}}$ as $4000$ and $4$, respectively, while $128$ and $4$ for the LDC-based models (i.e. LDC, KD-LDC and \ouralg). We opt for DeepConvNet for the Motor Imagery and ERN datasets, and EEGNet for the X11 and S4b benchmark, as the teacher model, respectively. The change point $P$ is $100$ in Motor Imagery and ERN, while $75$ for X11 and S4b. In curriculum, we set the easiest 65\% of data examples as the easy pool, the easiest 80\% as the \{easy + medium hard\} pool, while the easiest 95\% samples as the \{easy + medium hard + hardest\} pool for the Motor Imager and ERN datasets. In X11 and S4b, the percentages are $70\%$, $90\%$ and $100\%$.

\subsection{Main Results}
\begin{table*}[t]
    \small
    \setlength\intextsep{0pt}
    \setlength\lineskip{0pt}
    \captionsetup{width=0.95\linewidth}
  
    \caption{The \ouralg provides a better tradeoff between accuracy and inference efficiency compared to other methods. Note that models dominated by BMACs are typically more hardware-computation efficient than those dominated by FPMACs, as the BAMCs can be implemented in a massively parallel fashion in platforms like FPGA~\citep{zhang2020fracbnn}.}
    
    \centering
     \resizebox{0.92\textwidth}{!}{
    \setlength{\tabcolsep}{16pt}
    \begin{tabular}{cccccc}
    \toprule
   
    Dataset & Method & Accuracy & BMACs & FPMACs &  Model Size \\
     & & (\%) & ($\times 10^6$) & ($\times 10^6$) & (KB) \\
    \midrule
    \midrule
     \multirow{11}{*}{\textbf{Motor Imagery}} 
      & EEGNet~\cite{Lawhern2018} & \acc{85.33}{1.25} & 0 & 4.81 & 72.22\\
      & DeepConvNet~\cite{Lawhern2018} & \acc{92.83}{1.21}
      & 0 & 25.33 & 613.82\\
      & Binarized-DeepConvNet & \acc{57.68}{2.21} & 24.56 & 0.76 & 47.15\\
        & SVM (HALO)~\cite{hshalo} & \acc{58.42}{0.45} & 0 & $1.02 \times 10^{-3}$ & 4.60\\
        & MLP & \acc{86.12}{1.44} & 0 & 32.90 & 125.88\\
      & LeHDC & \acc{76.02}{0.44} & 4.06 & 0 & 527.81 \\
      \cmidrule(lr{1em}){2-6}
      & LDC & \acc{77.18}{0.89} & \multirow{5}{*}{\textbf{0.13}} & \multirow{5}{*}{\textbf{0}} & \multirow{5}{*}{\textbf{16.89}}\\
       & KD-LDC & \acc{77.89}{0.73} & \\
       & CTKD~\cite{li2022cur} w/ LDC & \acc{78.85}{0.62} & \\
    & \textbf{\ouralg} & \bacc{80.17}{0.83} & \\
      \midrule
      \midrule
       \multirow{10}{*}{\textbf{X11 and S4b}}
     
      & EEGNet~\cite{Lawhern2018}& \acc{80.04}{1.09} & 0 & 5.98 & 105.38\\
      & Binarized-EEGNet & \acc{56.14}{1.83} & 5.76 & 0.21 & 12.77\\
      & DeepConvNet & \acc{83.71}{2.07} & 0 & 28.99 & 892.01 \\
         & SVM (HALO)~\cite{hshalo} & \acc{53.55}{0.43} & 0 & $1.50 \times 10^{-3}$ & 6.01\\
      & LeHDC & \acc{68.64}{0.22} & 5.93 & 0 & 754.68\\
      \cmidrule(lr{1em}){2-6}
      & LDC & \acc{69.16}{1.04} & \multirow{5}{*}{\textbf{0.19}} & \multirow{5}{*}{\textbf{0}} & \multirow{5}{*}{\textbf{24.15}} \\
      & KD-LDC & \acc{69.68}{0.83} & \\
       & CTKD~\cite{li2022cur} w/ LDC  &  \acc{70.22}{0.64} & \\
    & \textbf{\ouralg} & \bacc{71.83}{0.77} & \\
    \midrule
    \midrule
       \multirow{10}{*}{\textbf{ERN}}
     
      & EEGNet~\cite{Lawhern2018} & \acc{82.84}{1.04} & 0 & 4.77 & 69.78\\
      & Binarized-EEGNet & \acc{59.63}{1.74} & 4.59 & 0.18 & 5.31 \\
      & DeepConvNet & \acc{86.67}{1.34} & 0 & 27.68 & 632.56\\
         & SVM (HALO)~\cite{hshalo} & \acc{55.80}{0.33} & 0 & $1.12 \times 10^{-3}$ & 4.38\\
      & LeHDC & \acc{72.63}{0.45} & 4.59 & 0 & 597.81 \\
       \cmidrule(lr{1em}){2-6}
      & LDC & \acc{73.34}{0.87} &  \multirow{5}{*}{\textbf{0.15}} & \multirow{5}{*}{\textbf{0}} & \multirow{5}{*}{\textbf{19.13}}\\
      & KD-LDC & \acc{73.86}{0.73}\\
       & CTKD~\cite{li2022cur} w/ LDC & \acc{74.42}{0.60}\\
    & \textbf{\ouralg} & \bacc{75.57}{0.62}\\
    \bottomrule
    \end{tabular}
    }
    \label{tab:res_base}
\vspace{-5pt}
\end{table*}

Table~\ref{tab:res_base} shows the main results of accuracy and efficiency across different methods among the datasets. 
From the accuracy perspective, the DeepConvNet and MLP have achieved the highest accuracy. However, their accuracy comes at the cost of heavy FPMAC operations, and requires over 40 times larger model size compared to the \ouralg.  Regarding the model efficiency, although the Binarized-DeepConvNet and LeHDC are also dominated by efficient BMACs operations, the \ouralg outperforms them by a large margin in terms of accuracy. As for the inference model size, SVM has the smallest number, but its accuracy is  \textasciitilde  20\% lower than the \ouralg on the evaluated tasks. In addition, the \ouralg consistently outperforms the other knowledge distillation methods like the plain KD LDC and CTKD w/ LDC across all evaluated EEG datasets.  In general, \ouralg has better balanced accuracy and efficiency on the EEG datasets compared to other methods. Note that methods like EEGNet show BMACs of $0$ as they only involve floating-point computations. Conversely, for methods like LDC, which only entail binary operations in inference, they have $0$ FMACs.

\subsection{Analysis of the \texorpdfstring{$\alpha$} Scheduler}
\begin{figure*}
    \centering
    \includegraphics[width=.95\textwidth]{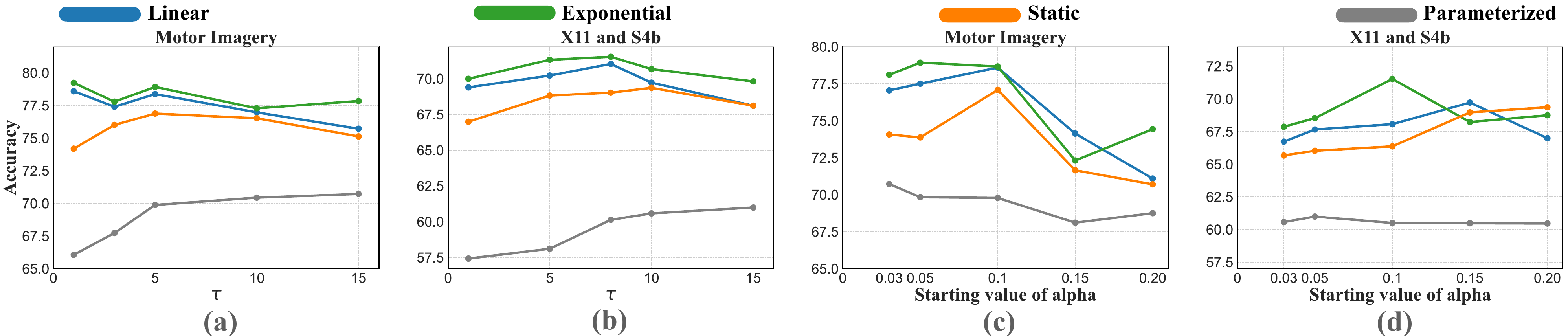}
      \vspace{-6pt}
      \captionsetup{width=\textwidth}
    \caption{Comparison of different $\alpha$ setups, \textit{without} employing data curriculum in the KD setting on the Motor Imagery, X11 and S4b datasets. (a), (b): With different temperatures $\tau$. (c), (d): With different starting values of $\alpha$. 
    }
    \label{fig:abl-alpha}
    \vspace{-10pt}
\end{figure*}

We show the results of using different $\alpha$ setups without data curriculum in the Figure~\ref{fig:abl-alpha}. The experiment results have demonstrated that utilizing an $\alpha$ scheduler to regulate the distillation level during the training process is more effective than using a static $\alpha$. It supports our belief that as the student model gains proficiency in tackling the task, it requires less knowledge distillation from a much more complex teacher model, in order to independently generate data representations using its own understanding based on the already good initialization.
Furthermore, as observed from the Figure~\ref{fig:abl-alpha} (a) and (b), the exponential $\alpha$ scheduler generally performs slightly better than the linear $\alpha$ scheduler across various temperatures. 
In contrast, using parameterizing $\alpha$ leads to the poorest accuracy. This is primarily due to the significant difference in magnitudes between the values of $\sL_{KD}$ and $\sL_{NLL}$ throughout the training process. Consequently, optimizing the parameter $\alpha$ tends to heavily favor one loss over the other, leading to suboptimal performance. 
We attribute the improvements to the smoother and gradual decay, as well as the fine-grained exploration during the early stage of the exponential change, as opposed to the linear one.

\subsection{Efficacy of Curriculum Data Order}
In the Table~\ref{tab:curricular}, we present results of using different data curriculum methods. Firstly, our experiments reveal that curriculum data order helps in the knowledge distillation setting. However, if not under the knowledge distillation setting, the curriculum training does \textit{not} significantly improve accuracy~\citep{wu2021curricula}, as demonstrated by comparing the results of LDC and Curri LDC. Second, our experiments show that anti-curriculum (as adopted in Anti-curri KD-LDC), which ranks and trains data from difficult to easy, adversely affects accuracy, lowering it by approximately $\sim$4\% compared to KD-LDC, which uses random data ordering, on the Motor Imagery dataset. Lastly, combining the curriculum data order (as in Curri KD-LDC) with the exponential $\alpha$ scheduler yields the best performance.

\begin{table}[t]
\small
    \setlength\intextsep{0pt}
    \setlength\lineskip{0pt}
    \captionsetup{width=\linewidth, skip=5pt}
    \caption{Efficacy of curriculum training on the Motor Imagery. 
    }
    \centering
    \resizebox{0.5\textwidth}{!}{
    \setlength{\tabcolsep}{6pt}
    \begin{tabular}{ccccc}
    \toprule
   
      & Static $\alpha$ & Linear $\alpha$  & \textbf{Expo $\alpha$} & param $\alpha$\\
    \midrule
    
    \textbf{Curri KD-LDC} & \bacc{78.04}{0.60} & \bacc{79.57}{0.86} & \bacc{80.17}{0.83} & \bacc{72.83}{0.56} \\
     KD-LDC & \acc{77.89}{0.73} & \acc{78.59}{0.64} & \acc{78.92}{0.57} &\acc{70.72}{0.47}\\
    Anti-curri KD-LDC & \acc{73.93}{0.81} & \acc{73.84}{0.74} & \acc{74.65}{0.70} & \acc{67.72}{0.83}\\

    \midrule
    \midrule
    \textbf{Curri LDC} & \multicolumn{3}{c}{\bacc{77.20}{0.57}}\\
    LDC & \multicolumn{3}{c}{\acc{77.18}{0.89}}\\
    Anti-curri LDC & \multicolumn{3}{c}{\acc{73.43}{1.20}}\\
    \bottomrule
    \end{tabular}
    }
    \label{tab:curricular}
\end{table}

In Table~\ref{tab:abla-loss} (a), we show the results of using loss-based rankings from the pretrained teacher model versus the pretrained student model. We observe that using the teacher's loss to order data results in significantly worse accuracy compared to using the student's loss. One plausible explanation could be that the teacher and student models have different perceptions of the difficulty of the same data examples. In fact, the Table~\ref{tab:abla-loss} (b) reveals a low overlap in rankings between teacher and student models, highlighting the need of scheduled learning to bridge the gap and reduce mismatch.

\begin{table}
\small
\caption{Data ordering analysis on the Motor Imagery dataset. (a) Accuracy comparison when data ordered by loss of teacher model vs. student model;  (b) Loss-based ranking intersection between teacher and student model. 
}
\scalebox{0.9}{
\setlength{\tabcolsep}{2pt}
\begin{tabular}{lcc}
    \toprule
     & \begin{tabular}{@{}c@{}}Order by \\ Teacher Loss\end{tabular} &  \begin{tabular}{@{}c@{}}\textbf{Order by} \\ \textbf{Student Loss}\end{tabular}\\
     \midrule
     ScheduledKD LDC & \acc{74.14}{0.66} & \bacc{80.17}{0.83}\\  
Curri LDC &  \acc{70.67}{0.58} & \bacc{77.20}{0.57}\\  
Curri LDC w/ KD & \acc{74.34}{0.61} & \bacc{78.04}{0.60}\\
    \bottomrule
    \multicolumn{3}{c}{(a)}
    \end{tabular}
    }
\hspace{2pt}
\scalebox{0.9}{
\setlength{\tabcolsep}{2pt}
\begin{tabular}{lc}
    \toprule
       &  \begin{tabular}{@{}c@{}} Overlapped \\ Rank (\%)\end{tabular} \\
    \midrule
    Hardest 30\% & 30.06\\
     Hardest 50\% & 50.25 \\
      Hardest 70\% &  70.72\\
    
    \bottomrule
    \multicolumn{2}{c}{(b)}
    \end{tabular}}
\label{tab:abla-loss}
\vspace{-10pt}
\end{table}

\section{Related Works}
In BCIs, real-time operation with minimal signal-acquisition-to-output delay is crucial, making computation complexity and model efficiency a focus in research~\citep{aggarwal2022review, medhi2022efficient, saichoo2022investigating, korkmaz2022efficient, belwafi2017hardware, kang2022meta, byun2022energy}.  The~\cite{bci-t} has reduced the calibration cost of brain signals by leveraging previously acquired EEG signals and projecting the new signals into a shared latent space. However, there has been a lack of measurements such as latency or model size to quantify their efficiency improvements. In \cite{bci-u}, they have proposed to decode and classify signal without storing them to enhance the response speed of BCI applications, but the models in their work are still computationally-expensive large deep nets.  
Using classic lightweight feature engineering models such as SVM in BCIs to meet latency or computation constraints can be a simple solution, but their accuracy can be unsatisfactory due to the limited computation capabilities~\cite{bci-e1, bci-u, hshalo}. 

Motivated by the observation that human brain operates on high dimensional data, the  HDC/VSA has emerged~\cite{kanerva2009hyperdimensional}. It offers a promising alternative for handling noisy time-series data such as brain signals, as it integrates learning capability and memory functions. Several recent studies have used the HDC/VSA in biosignal tasks for improved inference efficiency~\cite{bci-hdc1, bci-hdc2, bci-hdc3}. Despite a few attempts to enhance its accuracy, the HDC/VSA still lacks satisfactory performance~\cite{HDCSurveyReviewIEEECircuitMagazing20209107175, duan2022lehdc, imani2019quanthd, imani2019semihd}. The recently proposed LDC offers improved accuracy and efficiency, with a model size 100 times smaller during inference compared to HDC~\cite{duan2022braininspired, liu2023metaldc}. However, directly applying LDC to EEG datasets, known for strong noise, can yield unsatisfactory accuracy~\cite{bci-hdc1}. 

Knowledge distillation is a common practice to achieve model size compression and maintain accuracy in small architectures~\citep{hinton2015distilling}. The standard knowledge distillation has been shown to improve the performance of student models in various applications~\citep{kdapp1, kdapp2, kdapp3}. However, severe prediction distribution mismatch between teacher and student models can occur, especially when there is a large gap between their model capacities~\citep{cho2019efficacy,stanton2021does}. Existing works have proposed using intermediate features to transfer learned representations from the teacher to the student. However, this approach requires high computational cost and storage sizes~\citep{Zhao-2022-CVPR, heo2018knowledge, heo2019comprehensive, romero2015fitnets}. In comparison, our proposed method \ouralg introduces no additional intermediate between the teacher and student model. The work closest to us is~\cite{li2022cur}, where their proposed CTKD curriculums the distilling temperature. In contrast, our \ouralg directly adjusts the weights of soft targets to control the influence of the teacher's prediction, where the exponentially decreased $\alpha$ provides the student with more confidence to generate its own data representations as it progresses towards maturity. Our empirical evaluation demonstrates the \ouralg achieves higher accuracy than the CTKD on the EEG benchmarks.

\section{Discussion and Future Works}
In this work, we propose the \ouralg to 
regulate the distillation level by $\alpha$ and the order of the training data by curriculum, leading to improved knowledge transfer and accuracy on the evaluated EEG benchmarks. Instead of using classical feature engineering methods or DNNs, we opt for the LDC/VSA classifier for high efficiency with low inference computational cost and smaller memory footprint for tiny BCI devices demanding low latency.

\textbf{Limitation and Future Works}
We focus on the EEG-based BCI benchmarks, but it would be worthwhile to explore the use of other brain signals like invasive Electrocorticography (ECoG) and Functional Magnetic Resonance Imaging (fMRI) in future studies. While our proposed \ouralg has demonstrated improved accuracy compared to methods with similar model size (or MACs), it still falls short compared to modern DNNs in terms of accuracy. We therefore aim to explore more ways to bridge the accuracy gap between our method and DNNs. Additionally, we hope to further investigate on-chip power consumption measurements during inference for BCIs on realworld tiny devices.

\section{Acknowledgment}
Yejia Liu and Shaolei Ren were supported in part by the U.S. NSF grant CNS-2007115. Shjin Duan and Xiaolin Xu  were supported in part by the U.S. NSF grant CNS-2326597.

\clearpage
{
\bibliographystyle{plain}
\bibliography{ref}
}

\end{document}